\documentclass[letterpaper, 10 pt, conference]{ieeeconf}  

\IEEEoverridecommandlockouts                              
\usepackage{graphicx} 
\usepackage{amsmath} 
\usepackage{amssymb}  
\usepackage{multirow}
\usepackage{booktabs}
\usepackage[colorlinks,urlcolor=black,linkcolor=black,anchorcolor=black,citecolor=black]{hyperref}

\title{\LARGE \bf
TZC: Efficient Inter-Process Communication for Robotics Middleware with Partial Serialization
}

\author{Yu-Ping Wang$^{1}$, Wende Tan$^{1}$, Xu-Qiang Hu$^{1}$, Dinesh Manocha$^{2}$, Shi-Min Hu$^{1}$\\
\\
Video of applications can be found at \url{http://youtu.be/WLFitz26fqQ}.
\\
The source code can be found at \url{http://github.com/Jrdevil-Wang/tzc_transport}.
\thanks{$^{1}$Yu-Ping Wang, Wende Tan, Xu-Qiang Hu, and Shi-Min Hu are with the Department of Computer Science and Technology, Tsinghua University, Beijing, China. }%
\thanks{$^{2}$Dinesh Manocha is with the Department of Computer Science, University of Maryland, MD 20742, USA. }%
\thanks{Yu-Ping Wang is the corresponding author (wyp@tsinghua.edu.cn).}%
}

\begin{document}

\maketitle
\thispagestyle{empty}
\pagestyle{empty}

\begin{abstract}

Inter-process communication (IPC) is one of the core functions of modern robotics middleware.
We propose an efficient IPC technique called TZC (Towards Zero-Copy). As a core component of TZC, we design a novel algorithm called partial serialization. Our formulation can generate messages that can be divided into two parts. During message transmission, one part is transmitted through a socket and the other part uses shared memory. The part within shared memory is never copied or serialized during its lifetime.
We have integrated TZC with ROS and ROS2 and find that TZC can be easily combined with current open-source platforms. By using TZC, the overhead of IPC remains constant when the message size grows. In particular, when the message size is 4MB (less than the size of a full HD image), TZC can reduce the overhead of ROS IPC from tens of milliseconds to hundreds of microseconds and can reduce the overhead of ROS2 IPC from hundreds of milliseconds to less than 1 millisecond.
We also demonstrate the benefits of TZC by integrating it with TurtleBot2 to be used in autonomous driving scenarios. We show that by using TZC, the braking distance can be 16\% shorter than with ROS.

\end{abstract}

\section{Introduction}\label{sec:introduction}

Robotics software systems have complex architectures and are organized into modules. In modern operating systems, processes are used to provide isolation and resource management methods. Running each module separately is a common and reliable method for managing these modules. Meanwhile, modules must exchange information with each other. For example, a simultaneous localization and mapping (SLAM) module may obtain information from cameras, lasers, or sonar modules and provide calibrated maps to other modules such as visualization and navigation. Therefore, inter-process communication (IPC) is one of the core functions of robotics middleware.

Most robotics middleware, such as ROS~\cite{c4}, employs socket-based IPC methods~\cite{c5} because they provide a unified abstraction of whether different processes are running on the same computational device or on separate devices. However, the performance is not satisfying when transmitting large messages to multiple receivers (or subscribers). Maruyama et al.~\cite{c13} reported detailed performance evaluations for ROS and ROS2. They showed that the communication latency increases nearly linearly with the growth of the message size. To be specific, with data as large as 4MB, the median latency of ROS local communication was about 3ms and that of ROS2 local communication was about 10ms. Things become significantly worse when there are multiple subscribers. For the case of 1MB of data and 5 subscribers, the median latency of ROS2 was about 80ms. The performance of ROS2 is worse than that of ROS because ROS2 employs the DDS technique~\cite{c12} to provide the QoS feature, but the message formats of DDS and ROS2 are not unified and the translation procedure costs more time.

\begin{figure}[t]
\centering
\includegraphics[width=\columnwidth]{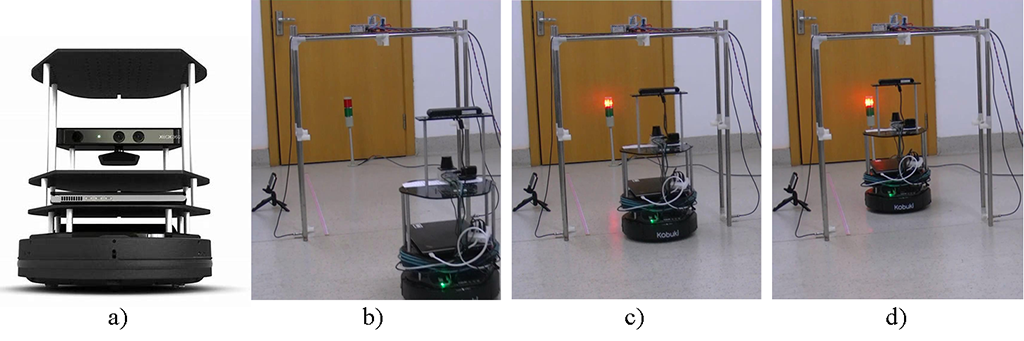}
\caption{Demonstration of our new IPC technique on TurtleBot2. TurtleBot2 (a) initially moves forward at 60cm/s (b). When it crosses the ``door,'' a red light is triggered (c). There are ROS nodes that constantly receive images from the HD camera node through IPC and detect signal lights. When the red light is recognized, the detecting node commands the TurtleBot to start braking (d). By using our TZC technique, the overall software latency is reduced from 200ms to 20ms and the braking distance is shortened by 10cm compared to ROS. The braking distance is an important metric for autonomous driving applications.}
\label{fig:teaser}
\end{figure}

The main causes of this high latency are copying and serialization operations. For example, if a robot is designed to help people find desired objects, images are captured by the camera module and provided to facial recognition and object recognition modules. When using ROS in this scenario, each \textit{Image} message captured by the camera node is (a) serialized into a buffer; (b) copied into the operating system kernel; (c) copied to the destination process; and (d) de-serialized into an \textit{Image} message. With the wide application of high-resolution cameras and LiDAR sensors, these operations result in high latency. The communication latency from the camera module to the other modules is over 20ms with ROS when the images are 1920x1080x24 bits. This latency will directly affect the response time and the user experience. For latency-sensitive robotics systems such as autonomous driving or real-time navigation in crowded scenarios, we need lower latency communication capabilities.

\begin{figure}[htb]
\centering
\includegraphics[width=\columnwidth]{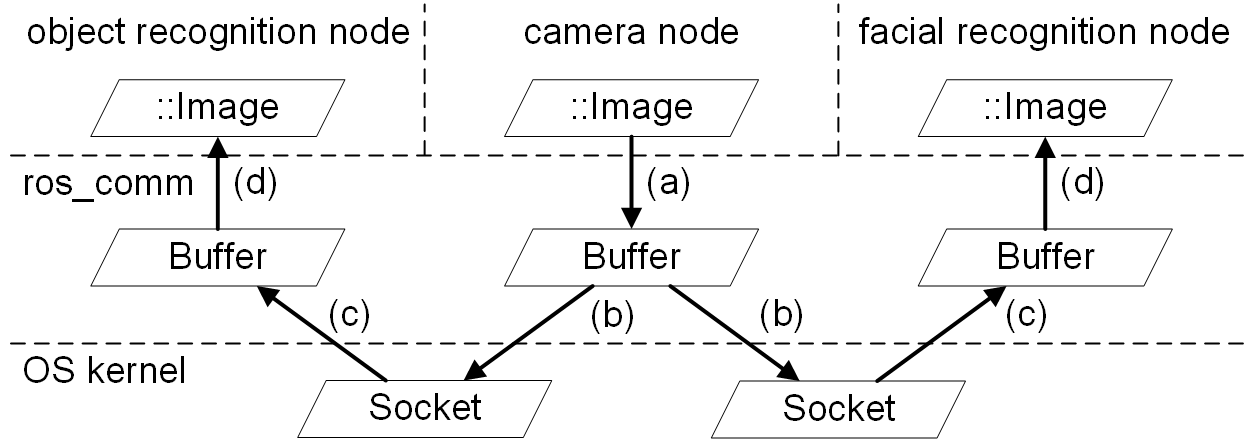}
\caption{The data flow of publishing images to two other ROS nodes. }
\label{fig:flow}
\end{figure}

\textbf{Main Results:} In this paper, we present an efficient IPC technique called TZC (Towards Zero-Copy) for modules that run on the same machine. As a core component of TZC, we design an algorithm called partial serialization. Our formulation can generate robotics message types that can be divided into two parts. During message transmission, one part is transmitted through a socket and the other part is stored within shared memory. Our partial serialization algorithm automatically separates data that are suitable for shared memory, which are also the majority of the message. As a result, TZC can provide the following benefits:

\begin{itemize}

\item \textbf{Zero-copied messages}. With the help of the partial serialization algorithm and a shared memory IPC method, most of the message is never copied or moved to other places during its lifetime. This feature causes the communication latency to remain constant when the message size grows, which is good for large message transmission.

\item \textbf{Publish-subscribe pattern}. This pattern is commonly used by robotics middleware and is proven to be scalable for scenarios with multiple senders and receivers (collective communication). TZC also employs this pattern and potentially enables the portability of robotic modules.

\item \textbf{Compatible synchronization}. By employing a socket-based method, TZC subscribers can be notified when a new message arrives by using compatible select/poll notification interfaces.

\item \textbf{Platform portability}. TZC is based on POSIX shared memory and socket and is therefore portable among POSIX-compatible operating systems such as Linux, macOS, VxWorks, QNX, etc.

\end{itemize}

We have integrated TZC with ROS and ROS2. Comparison results using benchmarks show that TZC can shorten the latency of ROS by two orders of magnitude while the state-of-the-art shared memory solutions~\cite{c18}~\cite{c16} can shorten the latency of ROS by less than one order of magnitude; and TZC can shorten the latency of ROS2 by three orders of magnitude and even better than the intra-process communication method of ROS2. We also demonstrate its benefits using robotics applications by porting it on TurtleBot2 to perform some tasks, shown in Figure~\ref{fig:teaser}. In this application, the TurtleBot moves forward at a uniform speed and starts braking when a red light is recognized by a 1080p camera. We observe that TZC shortens the braking distance by 16\% compared with ROS.

This paper is organized as follows. In Section~\ref{sec:relatedwork}, we give an overview of prior work on robotic systems and inter-process communication. In Section~\ref{sec:approach}, we introduce the design and implementation of the TZC framework in detail. Then we highlight the experimental results in Section~\ref{sec:evaluation}.

\section{Related Work}\label{sec:relatedwork}

\subsection{Latency in Robotic Systems}

Latency is a concern for many robotic applications. Many techniques have been proposed to reduce the latency in different modules of Micro Aerial Vehicles (MAV). Oleynikova et al.~\cite{c35} focus on obstacle avoidance and use FPGAs to achieve low latency. Xu et al.~\cite{c37} focus on 3D navigation and use Octrees to optimize the path planning module. Honegger et al.~\cite{c38} have developed a stereo camera system with FPGAs. Cizek et al.~\cite{c34} also aim to reduce the latency of the vision-based mobile robot navigation system. All these methods focus on reducing the latency of computer vision algorithms, either by software optimizations or hardware accelerations.

\subsection{IPC Methods in Robotics Middleware}

Most robotics middleware employs socket-based IPC methods~\cite{c5}. The socket interface is designed to communicate between processes, regardless of whether they are on the same machine or not~\cite{c1}. When processes are running on the same hardware, a NIC-level loopback is used without any network transmission. The common socket is a point-to-point style communication interface. To support collective communication efficiently, multi-path protocols are employed~\cite{c6}\cite{c8}\cite{c9}. Recently, with the development of ROS2, there have been big changes within the communication layer~\cite{c10}\cite{c11}. A similar problem is solved by employing QoS and DDS techniques~\cite{c12}. However, because ROS2 is still in development, there are still performance issues that need to be improved~\cite{c13}. Even with multi-path protocols, the transmitted buffer must be copied multiple times throughout middleware and kernel levels, which has an adverse effect on communication latency.

Shared memory methods provide promising solutions for eliminating copying operations. Shared memory allows two or more processes to share a given region of memory, which is the fastest form of IPC because the data does not need to be copied~\cite{c15}. Many message transport frameworks have been developed in~\cite{c18}, including a shared-memory-based transport. In this framework, a message is directly serialized into shared memory by the publisher and de-serialized from shared memory by the subscriber(s). As a result, these researchers report a speedup of about two-fold over ROS because the copying operations are eliminated but the serialization operations remain. Ach~\cite{c16} provides a communication channel by wrapping shared memory as a device file. It can solve the Head-of-Line Blocking problem, which makes it suitable for robotic scenarios. It is integrated to multiple robots such as Golem Krang~\cite{c31} and Aldebaran NAO~\cite{c32}. It is also employed by the RTROS project~\cite{c17}. We will compare our TZC framework with Ach in Section~\ref{sec:evaluation}.

In addition to the standardized IPC interface mentioned above, other methods have been designed for new kernel assistant IPC methods~\cite{c19}\cite{c20}\cite{c21}\cite{c22}\cite{c23}. Most of these methods are designed for high performance computing and emphasize throughput rather than latency. Moreover, these methods are not POSIX compatible and are therefore not commonly used by robotics middleware. 

We summarize the above methods with respect to the number of copying and serialization operations in Table~\ref{table:summary}. In the table, the numbers of copying and serialization operations are associated with the number of subscribers $sub$. As we can see, shared-memory-based middleware is generally better than socket-based middleware in this respect and TZC successfully reduced the number of these operations to zero.

\begin{table}[htb]
\caption{Summary of Robotics Middleware with respect to the Number of Copying and Serialization Operations.} \label{table:summary}
\centering
\begin{tabular}{|c|c|c|}
\hline
\multirow{2}{*}{IPC Method}  &   \multirow{2}{*}{Middleware}	& No. of copying and		\\
                        &                               & serialization operations	\\
\hline
\multirow{3}{*}{Socket} &   ROS~\cite{c4}            	& $1 + 3 \times sub$ \\
                        &   LCM~\cite{c6}            	& $2 + 2 \times sub$ \\
                        &   ROS2~\cite{c10}     		& $3 + 3 \times sub$ \\
\hline
Shared                  &   ETHZ-ASL~\cite{c18}         & $1 + 1 \times sub$ \\
Memory                  &   Ach~\cite{c16}              & $1 + 1 \times sub$ \\
\hline
Combined                &   TZC                         & $0 + 0 \times sub$ \\
\hline
\end{tabular}
\end{table}

\subsection{Serialization Methods}

Most of the fundamental frameworks above are designed for general buffer transmission. When applied to robotic scenarios, messages are complex structures that must be serialized into buffers before transmission.

Serialization is a traditional technique~\cite{c24} that transforms abstract data types into byte buffers for communication or persistent storage. Many frameworks and programming languages such as JSON~\cite{c25} and Protocol Buffers~\cite{c26} support automatic serialization. ROS and DDS employ their own serialization methods. Even for shared memory IPC, messages usually must be serialized before being copied to shared memory due to the usage of pointers. Libraries such as Boost~\cite{c27} help programmers handle pointers within shared memory. However, supporting subtype polymorphism still requires compile-time modifications~\cite{c28}.

Because serialization keeps all the information from the original message, the running time overhead of serialization is at least the same as that of the copy operation. As far as we know, for latency-sensitive applications, the only existing efficient solutions that can avoid serialization are intra-process communication methods such as ROS nodelet~\cite{c29} and ROS2 intra-process communication~\cite{c30}. Under these mechanisms, different modules run within the same process rather than in separate processes. Therefore, messages can be directly accessed by other modules without copying or serialization operations. However, the obvious drawback of intra-process communication is fault isolation. Since all modules run within the same process, when any module crashes, the entire system crashes. In Section~\ref{sec:evaluation}, we will show that, by eliminating the copy operation and employing partial serialization, our TZC technique performs comparable to inter-process communication.

\section{The TZC IPC Framework}\label{sec:approach}

\begin{figure}[htb]
\centering
\includegraphics[width=0.7\columnwidth]{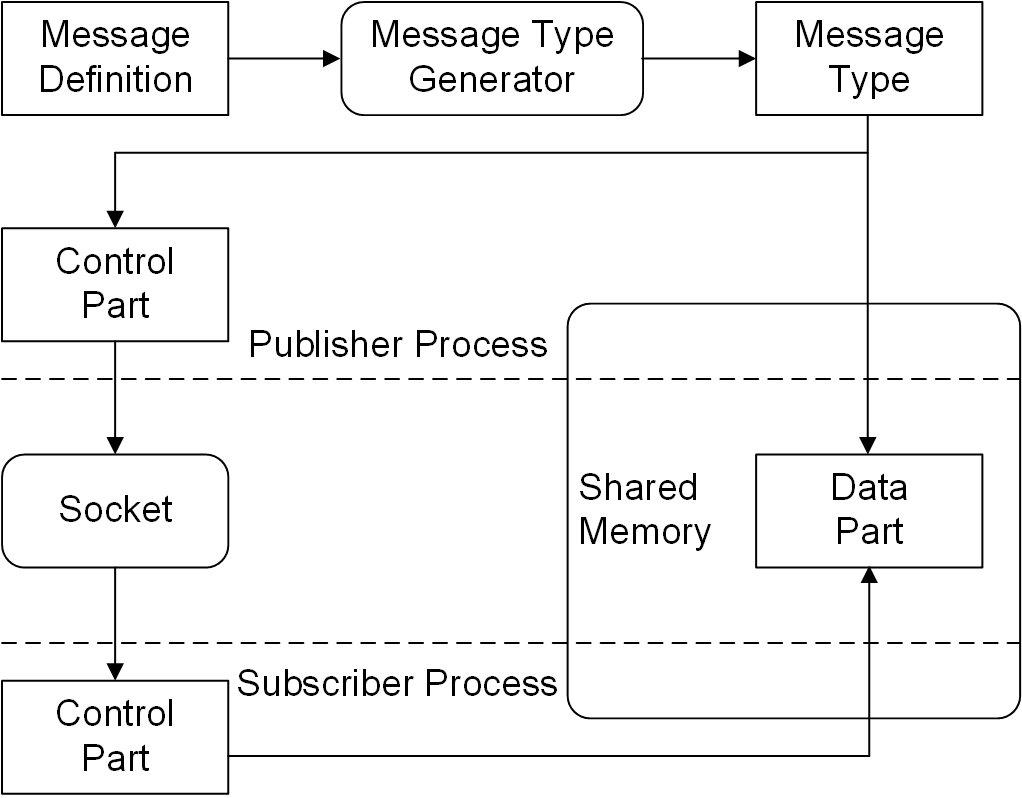}
\caption{The architecture of our TZC IPC framework. Each message is divided into two parts by the Message Type Generator. The control part is transmitted through the socket and the data part is shared through shared memory. Although the actual sizes of the control part and the data part depend on the message type and the actual message content, the control part is normally less than 4KB and the data part is probably more than 4MB. Thus, the communication latency is greatly reduced from common socket-based IPC frameworks.}
\label{fig:arch}
\end{figure}

In this section, we present our TZC IPC framework. In our method, we divide each message into two parts: the control part and the data part. The control part is relatively small and is transmitted through a socket-based IPC method after serialization. The data part is much larger and is shared through shared memory. Thus, the control part provides the compatible synchronization mechanism and the data part provides the zero-copy feature. Figure~\ref{fig:arch} shows the overall architecture of the TZC IPC framework.

The ROS communication framework can be seen as a special case of TZC in which the control part contains the whole message and the data part is empty. The whole message is serialized and transmitted through the socket. The ROS framework causes multiple copying and serialization operations. The ETHZ-ASL shared memory framework~\cite{c16} can also be seen as a special case of TZC in which the data part contains the whole message and the control part is a 32-bit handle. The ETHZ-ASL framework eliminates copying operations, but multiple serialization operations remain. This is because the whole message is a complicated structure that cannot be directly shared within shared memory without serialization. Therefore, the key of the TZC IPC framework is to distinguish which part belongs to the data part. We design a novel partial serialization algorithm to distinguish them.

\subsection{Partial Serialization Algorithm}

Our algorithm is a part of the Message Type Generator (in Figure~\ref{fig:arch}).  It has three main purposes. First, it decides which part of the message belongs to the control part. Second, it generates the serialization routine. Third, it organizes the data part.

Our main insight is that the memory structure of a fixed-length message is the same as that of the corresponding serialized message. Therefore, we can organize all variable-length arrays of fixed-length types within shared memory without serialization.

\begin{figure}[htb]
\centering
\includegraphics[width=0.8\columnwidth]{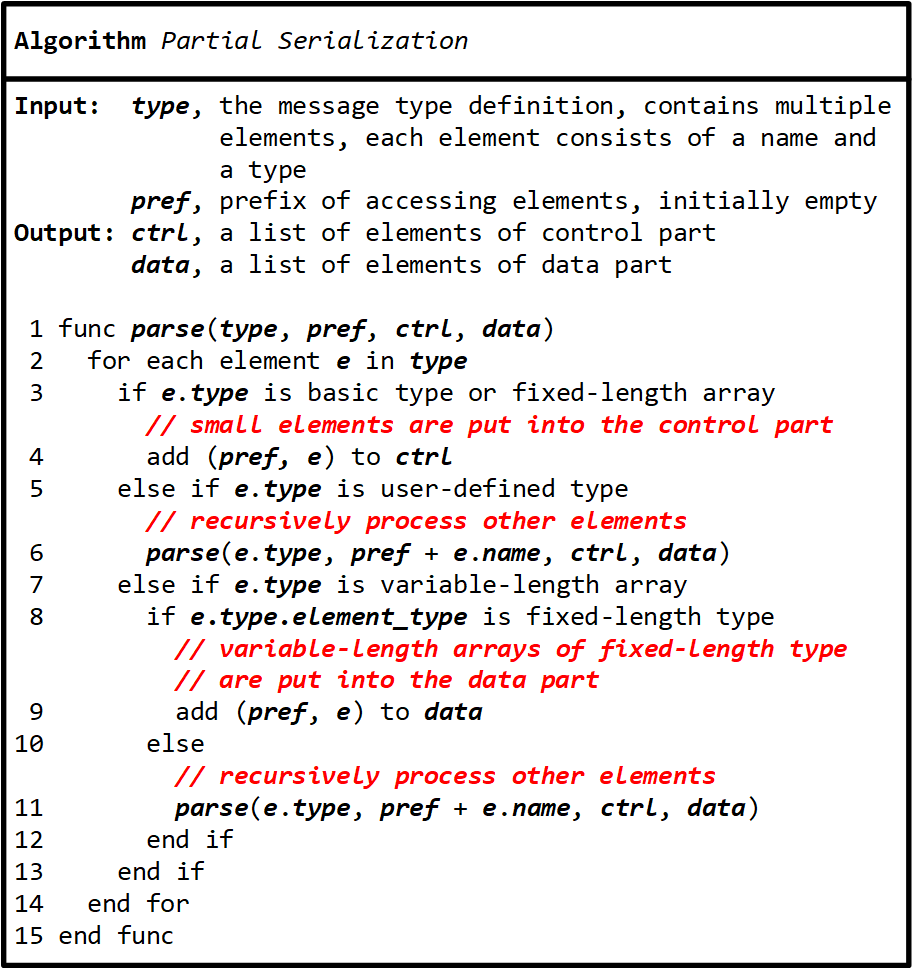}
\caption{The simplified partial serialization algorithm. Normal algorithm serializes all elements into one buffer. The partial serialization algorithm recursively classifies different elements into the control part (at line 4) or the data part (at line 9).}
\label{fig:algo}
\end{figure}

The partial serialization algorithm is shown in Figure~\ref{fig:algo}. It is a DFS-like algorithm, which recursively parses the message type definition tree. All variable-length arrays of fixed-length types are identified and classified into the data part and all other elements are classified into the control part.

\begin{figure}[htb]
\centering
\includegraphics[width=0.85\columnwidth]{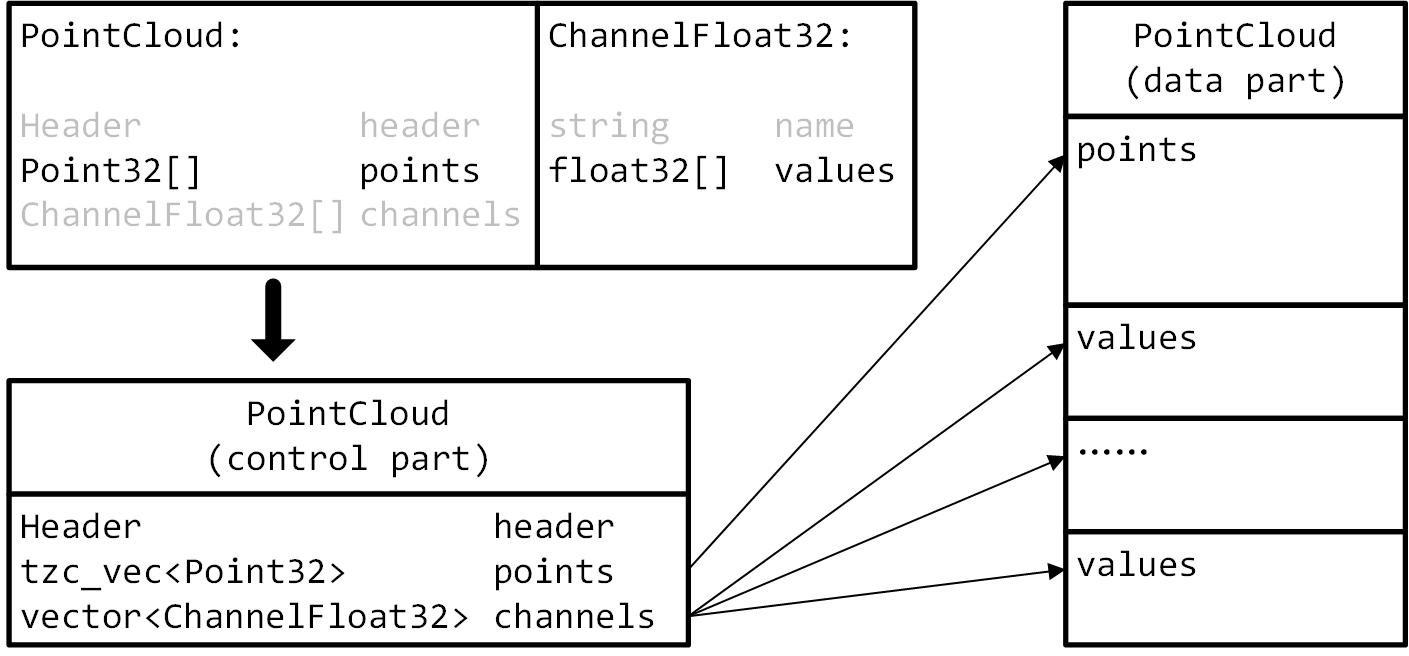}
\caption{An example of generating a point cloud message type using the ROS \textit{PointCloud.msg} definition. Gray elements belong to the control part and black elements belong to the data part. In the generated message type, some elements in the control part will point to the corresponding area within the data part. The data part is organized as one buffer within shared memory.}
\label{fig:msgs}
\end{figure}

For example, if our goal is to generate a point cloud message type using the ROS \textit{PointCloud.msg} definition shown in Figure~\ref{fig:msgs}, the \textit{header} element belongs to the control part because it is a variable-length type; the \textit{points} element belongs to the data part because it is a variable-length array of a fixed-length type \textit{Point32}; the \textit{channels} element belongs to the control part because it is a variable-length array of a variable-length type \textit{ChannelFloat32}; but inside each element of \textit{channels}, the \textit{values} element belongs to the data part. As a result, the \textit{PointCloud} message type contains a special vector \textit{points} of \textit{Point32}, which points to the \textit{points} area within the data part, and a vector \textit{channels} of \textit{ChannelFloat32}, which contains multiple \textit{values} that point to the corresponding \textit{values} area within the data part. In this way, the data part is organized as a single buffer within shared memory and the control part can access the message information in the same manner as accessing a ROS message.

After running the partial serialization algorithm, a serialization routine is generated. In this routine, elements in the \textit{ctrl} list are serialized sequentially and then the lengths of each of the elements in the \textit{data} list are serialized sequentially. Note that the actual content of the elements in the \textit{data} list does not participate in serialization. This is a key feature of our partial serialization algorithm. In addition, a memory organization routine is also generated. In this routine, the offset of each element is calculated with the length of each element in the \textit{data} list and these offsets are used to generate pointers in the control part that point to the corresponding area within the data part. These two routines are used within the TZC IPC framework and are not exposed to application developers. We will explain their usage in the next subsection.

Our algorithm still has one limitation. All basic types of ROS are fixed-length types, except for \textit{string}. Our algorithm handles all variable-length arrays of fixed-length types, but it cannot handle variable-length arrays of strings. However, this is not a serious issue in practice, as we will show in Section~\ref{sec:evaluation}.

\subsection{TZC Usage Pattern and Explanation}

\begin{figure}[htb]
\centering
\includegraphics[width=0.75\columnwidth]{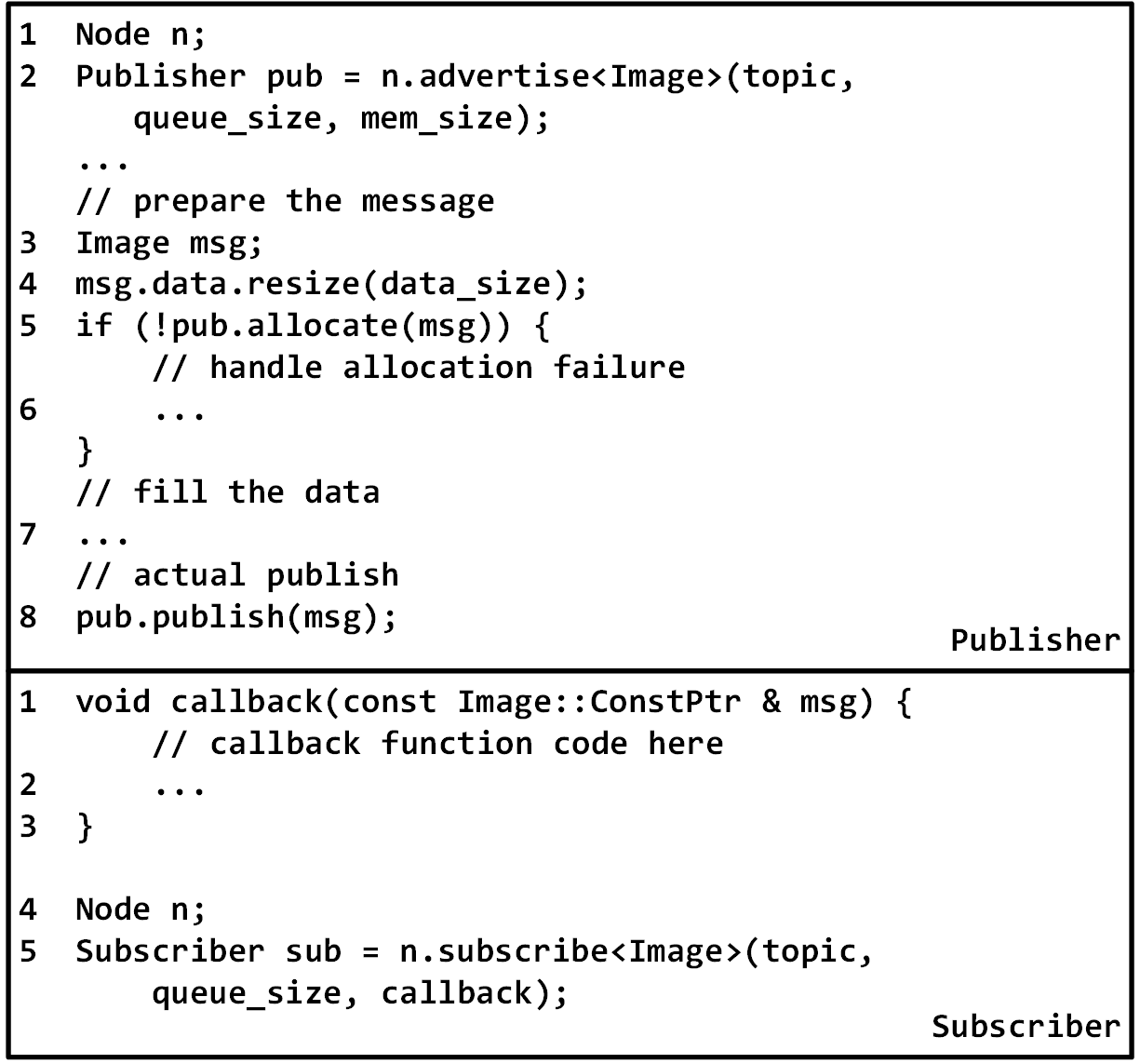}
\caption{A general TZC usage pattern. This pattern is very similar to the tutorial ROS node (see the footnote) and potentially enables the portability of robotic modules. Most improvements from TZC are beneath user interfaces. Thus, it is easy to integrate TZC to ROS-like middleware.}
\label{fig:usage}
\end{figure}

Figure~\ref{fig:usage} shows a general TZC usage pattern. It is very similar to the tutorial ROS node\footnote{\url{http://wiki.ros.org/ROS/Tutorials/WritingPublisherSubscriber\%28c\%2B\%2B\%29}}.
The concepts of \textit{Node}, \textit{Publisher}, \textit{Subscriber}, \textit{advertise}, \textit{subscribe}, and \textit{callback} are borrowed from ROS. The small difference visible on the interface is borrowed from the MPI~\cite{c2}. We would like to explain what happens under this usage pattern.

At the publisher side, when a \textit{Publisher} is created by the \textit{advertise} method, an associated shared memory region is created with respect to the newly added \textit{mem\_size} parameter, which decides the total size of the shared memory region. In this example, an \textit{Image} message is created (line 3) and published (line 8) in the same way as it would be in ROS. However, within the \textit{publish} method, the serialization routine generated by our partial serialization algorithm is called and only the control part is transmitted. Also note that at line 3, the newly created message only has the control part. The data part is created by calling a newly added \textit{allocate} method (line 5). The size of the data part is calculated with the associated data size, which must be assigned before allocation (line 4). If the allocation succeeds, the memory organization routine generated by the partial serialization algorithm is called to fill the associated pointers in the control part. Otherwise, if the allocation fails, we try to release some unused data parts of the older messages (see the next subsection). If this attempt also fails, the failure is reported to the application developer and the developer can handle this situation (line 6). On the subscriber side, the usage pattern is exactly the same as that of ROS. However, when a control part arrives, the memory organization routine generated by our partial serialization algorithm is called before the message is provided as an argument of the registered callback function (line 1). Within the callback function (line 2), the developer could perform any task needed to complete the desired functionality of the module, such as facial recognition or object recognition. During the whole message lifetime, the data part remains in the shared memory without any copying or serialization operations. This is the key that results in the reduced latency using TZC. We show the results in Section~\ref{sec:evaluation}.

\subsection{Modifying the Message Contents}

We should note that because the data part is shared within shared memory, the data part should not be modified by any subscribers. However, even if the developer needs to modify the data part, an additional copying operation is not always needed. For example, if a subscriber wants to filter the received image, instead of filtering it with an in-place algorithm, the developer can use a non-in-place algorithm that reads from the original image (in shared memory) and writes to a filtered image (in local memory). In most cases, the time of an in-place algorithm is almost the same as a non-in-place one. The main advantage of an in-place algorithm is saving memory. In addition, this is also an issue for all solutions that provide zero-copy features such as ROS nodelet.

\subsection{Message Buffering}

Since the size of the shared memory region is fixed after its creation, messages that are buffered in the shared memory region must be released at some point. Therefore, it is important that we manage the lifetime of messages. Our solution combines some commonly used techniques and we explain it below.

To manage messages across their lifetimes, we add a reference count, a magic number, and a double-linked node at the head of each data part. The reference count indicates how many times the data part is referenced across the whole system. This number is the same as the number of associated control parts. Therefore, we can maintain the reference count whenever a control part is constructed or destroyed. We adapt a laziness strategy that does nothing when the reference count hits zero; a data part can only be released when a publisher fails to allocate another data part. We never release a data part when it is referenced by any control parts. However, there is still a chance that the data part is released before a subscriber receives the control part. The magic number is used to valid the data part when a subscriber receives a control part. Finally, the double-linked node is used to help find other data parts within the same shared memory region. These data structures are not exposed to developers to avoid corruption.


\subsection{Publication Policies}

When an allocation failure occurs in the \textit{allocate} method, there are several reasonable ways to handle it that show different publication policies.

1) \textit{Best effort}. Under this policy, the \textit{allocate} method will repeatedly release non-referenced data parts either until there is enough memory space or until nothing else can be released. New messages are always prior to old messages, which may provide better real-time features. This is also one of the design principles of Ach. This policy is suitable for most scenarios when older messages have lower value.

2) \textit{Worst effort}. Under this policy, if the \textit{allocate} method finds that an old message is still in use, the new message is discarded. This policy is designed so that if an old message is being used, the following newer messages will be used in the future even if the reference count is currently 0. This policy is suitable for maintaining the continuity of messages as long as possible, such as for audio messages.

3) \textit{Medium effort}. Under this policy, the \textit{allocate} method will try to release non-referenced data parts several times. The maximum number of attempts is a pre-defined value. This value should not be 0, because this is the only time when TZC really releases the data part of a message. This policy is a compromise of 1) and 2).

4) \textit{Blocking}. The allocate method will be blocked until the oldest message is not in use. This policy should not be implemented, because it will cause severe synchronization issues and potentially considerable latency.

Our current implementation follows the \textit{Medium effort} policy, but it is easy to change the policy to adapt to actual situations.

\section{Evaluation}\label{sec:evaluation}

In this section, we show our benchmark results and applications of TZC. The benchmark experiments are performed on a Lenovo ThinkPad E440 laptop with Intel Core i5-4200M @2.5GHz processor (2 cores, 4 threads) and 8GB memory. The applications are built on the commonly used TurtleBot2 with the same laptop as the computational device and an ASUS Xtion PRO as the input sensor. The ROS version is Indigo on Ubuntu 14.04.

\subsection{Benchmark Results}\label{sec:benchmark}

Two key factors significantly affect the communication latency, i.e. the message size and the number of subscribers. For clarity, we generate 8 test cases by changing the message size from 4KB to 4MB and the number of subscribers from 1 to 8. For each test case, the publisher generates a message, records a time stamp in the message, and publishes the message to the subscriber(s). Each subscriber will record the latency and discard the message after receiving each message. This procedure is repeated 1000 times at 30 Hz.

\begin{figure}[htb]
\centering
\includegraphics[width=0.9\columnwidth]{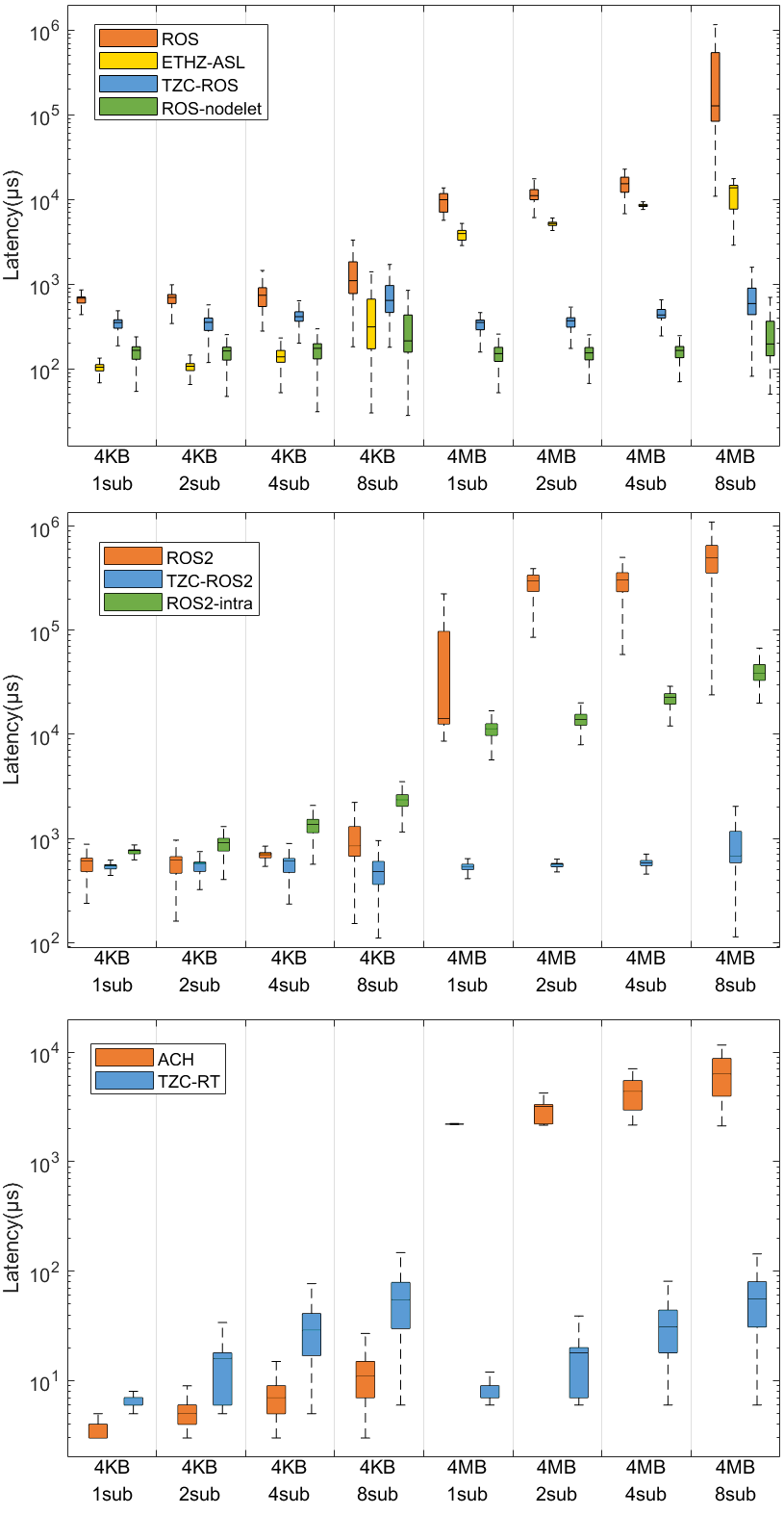}
\caption{Performance results on benchmarks. Each subscriber will record the latency and discard the message after receiving each message. In the first group, TZC-ROS (blue) outperforms ROS (orange) and ETHZ-ASL (yellow) by up to two orders of magnitude. In the second group, TZC-ROS2 (blue) outperforms ROS2 (orange) by up to three orders of magnitude. In the third group, TZC-RT (blue) outperforms Ach (orange) by up to two orders of magnitude. In addition, TZC-ROS and TZC-ROS2 show results comparable to intra-process communication methods (green).}
\label{fig:performance}
\end{figure}

For better comparison, we organize the results into three groups of box plot figures, shown in Figure~\ref{fig:performance}. The horizontal axis shows a different test case and the vertical axis shows the communication latency in the logarithmic coordinates.

In the first group, we integrate TZC with ROS to produce TZC-ROS. As we can see, the latency of ROS increases when the message size grows and the number of subscribers increases. In the test case involving a 4MB message and 8 subscribers, the mean latency of TZC-ROS is still under 1ms, which outperforms ROS by two orders of magnitude. In contrast, ETHZ-ASL only outperforms ROS by less than one order of magnitude.

In the second group, we integrate TZC with ROS2 to produce TZC-ROS2. As we can see, the latency of ROS2 becomes higher when the message size grows and remains stable when the number of subscribers increases. In the test case involving a 4MB message and 8 subscribers, the mean latency of TZC-ROS2 outperforms ROS2 by three orders of magnitude.

In the third group, we integrate TZC with a Linux / Xenomai / RtNet real-time platform to produce TZC-RT. As we can see, the latency of Ach becomes higher when the message size grows and the number of subscribers increases. When the message size is small, the mean latency of Ach outperforms TZC-RT by less than one order of magnitude; when the message size is large, the mean latency of TZC-RT outperforms Ach by up to two orders of magnitude.

\subsection{Performance Comparison}\label{sec:performance}

\textbf{TZC vs. Other shared memory solutions.} From Figure~\ref{fig:performance}, we can see that TZC can outperform the state-of-the-art solutions. As we have claimed in Table~\ref{table:summary}, although ETHZ-ASL and Ach employs shared memory, they still need serializations and cannot provide the zero-copy feature. The serialization routine takes as much time as the copying operation. As a result, the state-of-the-art shared memory solutions can only outperform ROS by less than one order of magnitude. These results are consistent with the results in the ETHZ-ASL ROS Wiki page and the Ach paper. However, TZC can provide the zero-copy feature by employing the partial serialization algorithm. Therefore, the message size no longer affects the latency after using TZC.
 
\textbf{TZC vs. Other socket-based solutions.} Except for ROS and ROS2, there are many other IPC frameworks such as DDS and MPI that are not designed for robotic software. They need to be properly encapsulated before used as robotic middleware. For example, the communication layer of ROS2 employs DDS. From Figure~\ref{fig:performance}, we can see that ROS2 performs more stable than ROS when the message size is smaller. This is the benefit of employing the DDS middleware. However, DDS also brings more latency when the message size becomes larger. This is because the DDS middleware has its own serialization methods and the transformation between a ROS2 message and a DDS message brings more latency. Overall, although we have not compared TZC with DDS directly, we have compared it with DDS through ROS2.

\textbf{TZC vs. Intra-process communication.} We also test the performance of the ROS nodelet and ROS2 intra-process communication. The mean latency of TZC-ROS is of the same order of magnitude as that of the ROS nodelet, but the ROS nodelet performs more stably when the number of subscribers increases. This is because TZC still employs the socket-based mechanism for transmitting the control part of the message, which gets worse when the subscriber number increases. On the other hand, ROS nodelet is a full zero-copy solution, but its main limitation is running all modules in the same process, which decreases reliability. Surprisingly, TZC-ROS2 can outperform ROS2 intra-process communication. Maruyama et al.~\cite{c13} reported similar results, indicating that the ROS2 intra-process communication needs to be improved.

\begin{figure}[htb]
\centering
\includegraphics[width=0.9\columnwidth]{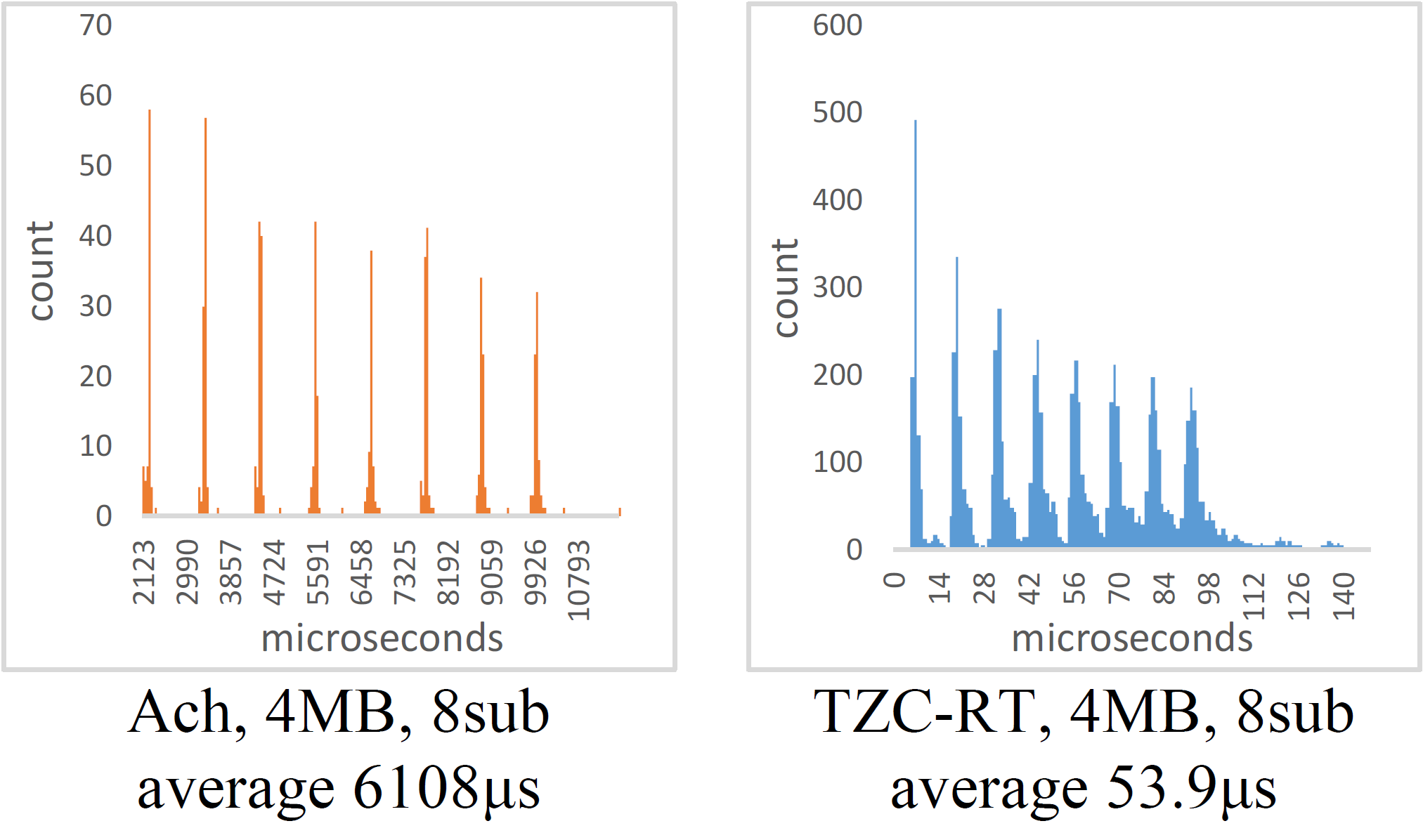}
\caption{Two special performance results on benchmarks shown with histograms. Both of them are for the test case of 4MB messages and 8 subscribers. The results show obvious periodicity, because multiple subscribers are nearly working in sequence on real-time system. The period among the peaks is approximately the execution time for a single subscriber. The period for Ach (orange) is about 1.1ms, which is mainly caused by serializing the 4MB messages; the period for TZC-RT (blue) is about 12us.}
\label{fig:performance-rt}
\end{figure}

\textbf{TZC for real-time}. Furthermore, we show two special results with histograms in Figure~\ref{fig:performance-rt}. The result for Ach is shown in orange and the result for TZC-RT is shown in blue. Both of them are for the test case of 4MB messages and 8 subscribers. As we can see, the results show obvious periodicity, because multiple subscribers are nearly working in sequence on real-time system, since there are no other tasks that could preempt their executions. Therefore, the period among the peaks is approximately the execution time for a single subscriber. The period for Ach is about 1.1ms which is mainly caused by copying the 4MB messages; the period for TZC-RT is about 12.3us which indicate that the messages are never copied or serialized.

\subsection{Reliability}

To compare the reliability of different methods, we also count the number of messages received by each subscriber during the performance experiment.

\begin{figure}[htb]
\centering
\includegraphics[width=0.85\columnwidth]{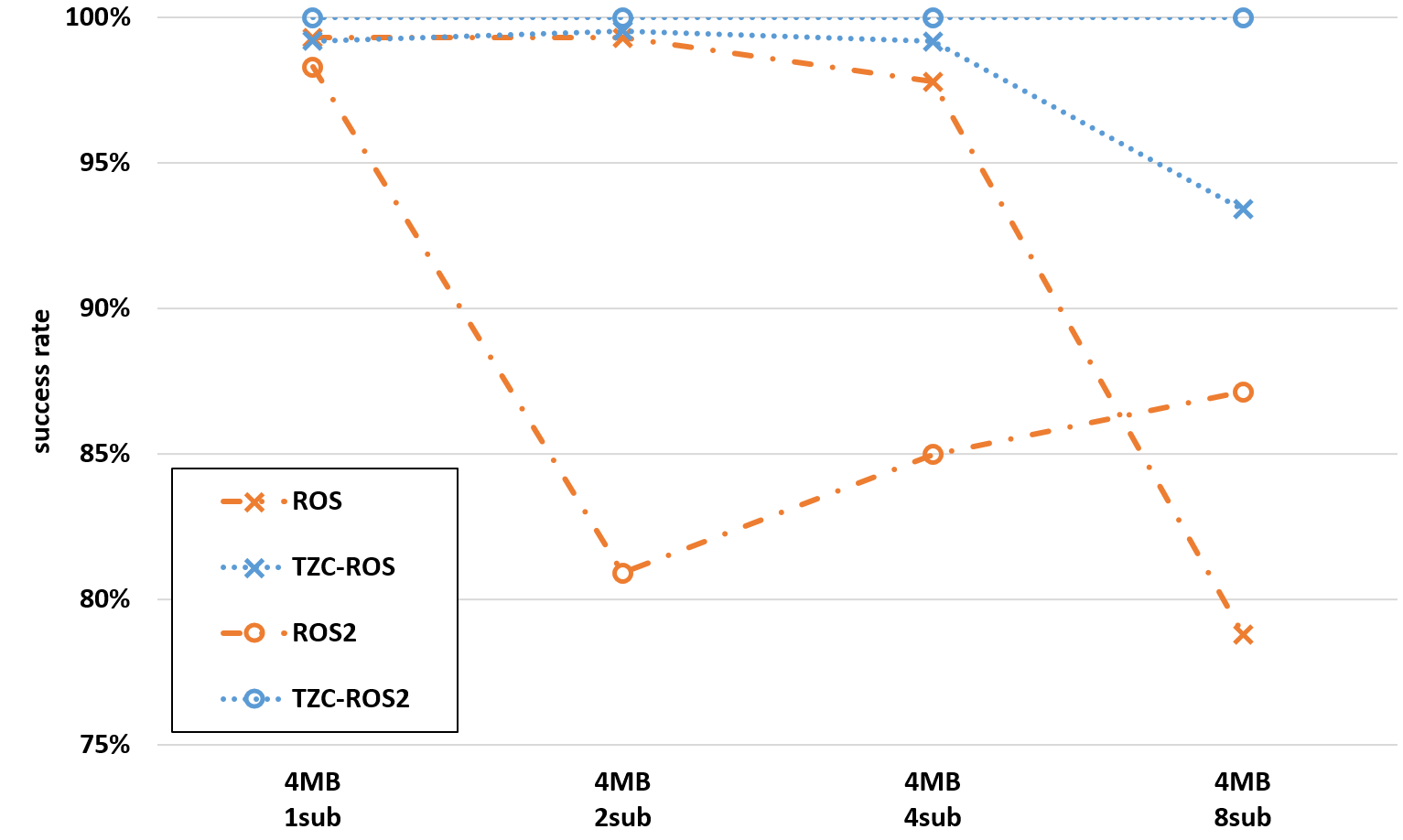}
\caption{Reliability results on benchmarks. As CPUs spend less time copying and serializing messages when using TZC, the probability of successful communication is increased. Particularly when using TZC-ROS2, subscribers can always receive all messages. }
\label{fig:reliability}
\end{figure}

Figure~\ref{fig:reliability} shows some of the results. The results from the real-time group are omitted because all messages are successfully received with Ach and TZC-RT. As we can see, as the number of subscribers increases, more messages are lost no matter which method is used. However, the success rates are still relatively high using our TZC framework. This is because CPUs spend less time copying and serializing messages when using TZC, which may lead to a higher possibility of successful communication. This is the key idea of the TZC framework. Particularly when using TZC-ROS2, subscribers can always receive all messages. This is because the default QoS configuration of DDS makes ROS2 more reliable than ROS when transmitting small messages. Interestingly, the success rate of ROS2 is lower for two subscribers than four. We cannot explain this result, but we guess it is associated with the DDS QoS policy.

\subsection{Compatibility}

TZC is easily integrable with current ROS middleware, which is reflected in two aspects of the process.

First, nearly all ROS message types can be supported by TZC and a significant number of them result in reduced latency. Six commonly used ROS meta-packages relating to messages are tested, including \textit{common\_msgs}, \textit{image\_transport\_plugins}, \textit{neonavigation\_msgs}, \textit{tuw\_msgs}, \textit{astuff\_sensor\_msgs}, and \textit{automotive\_autonomy\_msgs}. The supported result is shown in Table~\ref{table:typesupport}. As we can see, there are 358 message types within these packages. Among them, 104 of them contain variable-length arrays, which are potentially large messages and need TZC support. 98 of them can be supported, which means at least one of the variable-length arrays is of a fixed-length type. Among the 6 unsupported message types, 3 of them use strings as identifications or debugging information and the other 3 are associated with \textit{KeyValue}.

\begin{table}[htb]
\caption{Most ROS message types are supported by TZC.} \label{table:typesupport}
\centering
\begin{tabular}{|c|c|c|c|c|}
\hline
\multirow{2}{*}{}	& \multirow{2}{*}{Total} 	& Need		& \multirow{2}{*}{Supported}	& Not \\
					&							& Supported	&								& Supported \\
\hline
Number				& 358 		& 104		& 98		& 6 \\
\hline
Percent (/Total)	& 100\% 	& 29.1\%	& 27.4\%	& 1.6\% \\
\hline
Percent (/Need)		& - 		& 100\%		& 94.2\%	& 5.8\% \\
\hline
\end{tabular}
\end{table}

Second, only a minor part of the code needs to be changed when applying TZC. We have applied TZC to several ROS packages, mainly those associated with images and point clouds, and fewer than 100 lines of code need to be changed.

\subsection{Applications}

To show the improvements in robotic performance because of reduced latency, we highlight the performance on two latency-sensitive applications.

At the first application shown in Figure~\ref{fig:teaser}, the TurtleBot initially moves forward at a uniform speed. During this time, images are periodically captured and published to four different nodes. The first node logs the time stamps of the received images. The other three nodes try to recognize whether a red, green, or yellow light has been lit. Once there is a lit red light recognized, the node will instruct the TurtleBot to stop moving forward. For better comparison, the red light is turned on by a pair of laser switches that are triggered when the TurtleBot crosses a certain line.

\begin{figure}[htb]
\centering
\includegraphics[width=0.5\columnwidth]{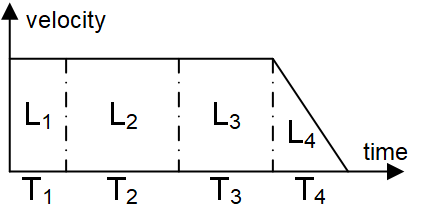}
\caption{The brake distance $L$ can be divided into four parts. Among them, $L_2$ directly reflects the middleware communication latency and the other parts are approximately constant under our experimental setup. Therefore, the difference in $L$ reflects the reduced latency.}
\label{fig:velocity}
\end{figure}

We record the brake distance $L$, which is measured between the stop position and the line that triggers the laser switch. This distance can be divided into four parts, shown in Figure~\ref{fig:velocity}. The first part $L_1$ is caused when the latency $T_1$ from the laser switch is triggered and the red light is turned on. The second part $L_2$ is caused when the latency $T_2$ from the red light is turned on and the recognition node receives the image. The third part $L_3$ is caused by the recognition algorithm. The fourth part $L_4$ is caused by the TurtleBot's deceleration process. Among these parts, $L_2$ directly reflects the middleware communication latency and the other parts are approximately constant. Therefore, the difference in $L$ reflects the reduced latency.

The comparison result shows that the difference in $L$ is about 10 cm with (53 cm) or without (63 cm) the TZC framework. The application logs show that $T_2$ is reduced from about 160ms to about 1ms after using TZC; $T_3$ is always about 20ms; $T_1$ and $T_4$ are hardware latency which cannot be measured by software.

We should emphasize that there are several parameters that impact the results. The workloads of the above applications are much heavier than those used in the performance tests. In this case, messages are more likely to be lost if new messages keep coming while the CPUs are busy transmitting messages or processing old messages. We use the \textit{matchTemplate / minMaxLoc} function of the OpenCV library to accomplish the light recognition. It is not very fast in dealing with a 1920x1080 image. Therefore, we set a region of interest (ROI) at the center of the image. A smaller size for the ROI leads to faster recognition, which costs less CPU time and means better reliability; however, a smaller size of the ROI also leads to a shorter period when the red light appears in the ROI, which means fewer key images will miss the opportunity for recognition. We have run the applications dozens of times with different parameters and the results in the video are chosen to show our contribution more clearly. Nevertheless, the results of TZC are generally better than those of ROS.

Furthermore, it is easy to translate this application to automatic driving or navigation scenarios. Analogically, when an automobile moves at 60 km/h (which is about 27.8 times the speed of the TurtleBot in the experiment), the same reduced latency causes the difference of the brake distance to be 2.78 meters.

\begin{figure}[htb]
\centering
\includegraphics[width=0.75\columnwidth]{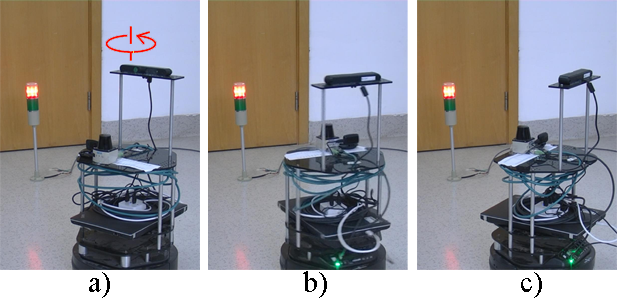}
\caption{Application of TZC to improve task performance. The TurtleBot2 rotates counterclockwise in situ (a). When a red light is captured by the camera and recognized by the ROS node (b), it commands the TurtleBot to start braking (c). When using ROS, many images are lost by ROS IPC and the TurtleBot rotates once more in many test rounds. By using TZC, the image transmission is more reliable and the TurtleBot stops more quickly. }
\label{fig:exp2}
\end{figure}

We highlight the second application in Figure~\ref{fig:exp2}. A similar experimental principle to that in the first application is used. The TurtleBot rotates counterclockwise in situ. Once there is a lit red light recognized, the node will inform the TurtleBot to stop rotation. The result shows that when using TZC, the TurtleBot stops about $30^\circ$ after the red light, but when using ROS, the TurtleBot rotates once more and stops about $60^\circ$ after the red light. This is because of the reliability issue. With this practical workload, CPUs are busier than the benchmarks and the image with a lit red light is easily discarded by ROS since more new messages have arrived. The result shows that TZC is more reliable and efficient.

\section{Conclusion and Future Work}\label{sec:conclusion}

We have presented an improved IPC framework called TZC to reduce the latency. By designing a novel partial serialization algorithm and combining the advantages of the socket-based and the shared memory IPC methods, the overall communication latency remains constant when the message size grows.

We have integrated TZC with ROS and ROS2. The benchmark evaluation shows that TZC can reduce the overhead of IPC for large messages by two or three orders of magnitude. We also highlight the performance on two latency-sensitive applications with TurtleBot2. Due to the improved efficiency and reliability from TZC, the performances of the applications are obviously improved. Thus, we show that TZC provides an efficient IPC framework for latency-sensitive applications.

Currently, our TZC framework only supports IPC on the same machine. To simultaneously enable communication among different machines, we could make use of the publisher-subscriber link to realize this feature. This is our future work. Another future work is to automatically transform a ROS or ROS2 node into a TZC node. It is promising by using program analysis techniques.

Besides, with the wide application of the General-Purpose GPU technique, much time is spent copying data between CPUs and GPUs. Therefore, the zero-copy technique has received more attention~\cite{c41}\cite{c42}. It would be interesting to inspect the robotic software with respect to the performance on heterogeneous hardware.

\addtolength{\textheight}{-8.5cm}

\end{document}